\pdfoutput=1
\documentclass{article}

\usepackage{arxiv}
\usepackage{amsmath}
\usepackage[dvipsnames]{xcolor}
\usepackage[utf8]{inputenc} 
\usepackage[T1]{fontenc}    
\usepackage{hyperref}       
\usepackage{url}            
\usepackage{booktabs}       
\usepackage{tabularx}
\usepackage{amsfonts}       
\usepackage{nicefrac}       
\usepackage{microtype}      
\usepackage{lipsum}		
\usepackage{graphicx}
\graphicspath{{./contents/images/}}

\usepackage{doi}

\usepackage{enumitem}

\usepackage{mfirstuc} 
\usepackage{stringstrings} 

\usepackage{schemata}
\usepackage{tikz}
\usepackage{tikz-qtree}

\usepackage{multirow}
\usepackage{graphicx}
\usepackage{caption}
\usetikzlibrary{trees,decorations.text,math,calc, positioning, arrows.meta}

\date{}

\renewcommand{\shorttitle}{\textit{arXiv} Template}

\begin{document}


\newcommand{\eg}{\textit{e.g.,~}}
\newcommand{\etc}{\textit{etc.}}

\newcommand{\comment}[1]{\textcolor{ForestGreen}{\textit{Comment: #1}}}  
\newcommand{\outline}[1]{\textcolor{Blue}{\textit{\textbf{[#1]}}}}

\newcommand{\textquote}[1]{\textcolor{Gray}{#1}}
\newcommand{\yyq}[1]{\comment{#1}}  

\newcommand{\app}{data augmented LLM applications}

\newcommand{\firstlevel}{explicit fact queries}
\newcommand{\secondlevel}{implicit fact queries}
\newcommand{\thirdlevel}{interpretable rationale queries}
\newcommand{\forthlevel}{hidden rationale queries}

\title{\Large \bf Retrieval Augmented Generation (RAG) and Beyond: A Comprehensive Survey on How to Make your LLMs use External Data More Wisely}

\author{ 	
    Siyun Zhao ,  Yuqing Yang , Zilong Wang,  Zhiyuan He,  Luna K. Qiu,  Lili Qiu\\
    Microsoft Research Asia\\
	\texttt{\{siyunzhao,yuqing.yang,wangzilong,zhiyuan.he,lunaqiu,liliqiu\}@microsoft.com} \\
}
\maketitle

\begin{abstract}
Large language models (LLMs) augmented with external data have demonstrated remarkable capabilities in completing real-world tasks. External data not only bolsters the models' domain-specific expertise and temporal relevance but also diminishes incidences of hallucination, thereby enhancing both the controllability and interpretability of outputs. Techniques for integrating external data into LLMs, such as Retrieval-Augmented Generation (RAG) and fine-tuning, are gaining increasing attention and widespread application. Nonetheless, the effective deployment of data-augmented LLMs across various specialized fields presents substantial challenges. These challenges encompass a wide range of issues, from retrieving relevant data and accurately interpreting user intent to fully harnessing the reasoning capabilities of LLMs for complex tasks. We believe that there is no one-size-fits-all solution for data-augmented LLM applications. In practice, underperformance often arises from a failure to correctly identify the core focus of a task or because the task inherently requires a blend of multiple capabilities that must be disentangled for better resolution. In this survey, we propose a RAG task categorization method, classifying user queries into four levels based on the type of external data required and the task’s primary focus: explicit fact queries, implicit fact queries, interpretable rationale queries, and hidden rationale queries. We define these levels of queries, provide relevant datasets, and summarize the key challenges and most effective techniques for addressing these challenges. Finally, we discuss three main forms of integrating external data into LLMs: context, small model, and fine-tuning, highlighting their respective strengths, limitations, and the types of problems they are suited to solve. This work aims to help readers thoroughly understand and decompose the data requirements and key bottlenecks in building LLM applications, offering solutions to the different challenges and serving as a guide to systematically developing such applications.
\end{abstract}

\section{Introduction}

Large Language Models (LLMs) have demonstrated remarkable capabilities, including extensive world knowledge and sophisticated reasoning skills. Despite these advancements, there are significant challenges in effectively deploying them across various specialized fields. These challenges include issues like model hallucinations, misalignment with domain-specific knowledge, among others. Incorporating domain-specific data, particularly private or on-premise data that could not be included in their initial training corpus, is crucial for tailoring LLM applications to meet specific industry needs. 
Through techniques like RAG and fine tuning, \app{} have demonstrated advantages over applications built solely on generic LLMs, in several aspects:

\begin{itemize}
    \item \textbf{Enhanced Professionalism and Timeliness:} The data used to train LLMs often lags in timeliness and may not cover all domains comprehensively, especially proprietary data owned by users. \MFUsentencecase{\app} address this issue by providing more detailed and accurate answers for complex questions, allowing for data updates and customization.
    
    \item \textbf{Alignment with Domain Experts:} Through the use of and learning from domain-specific data, \app{} can exhibit capabilities more like domain experts, such as doctors and lawyers.
    
    \item \textbf{Reduction in Model Hallucination:} \MFUsentencecase{\app} generate responses based on real data, grounding their reactions in facts and significantly minimizing the possibility of hallucinations.
    
    \item \textbf{Improved Controllability and Explainability:} The data used can serve as a reference for the model's predictions, enhancing both controllability and explainability.
\end{itemize}

Despite the enthusiasm for these advancements, developers often struggle and have to invest a significant amount of human labor to meet its expectations (\eg achieving a high success rate in question answering). Numerous studies~\cite{nie2024survey, huang2024comprehensive, yang2024large, ahn2024large, wang2024large} highlight the challenges and frustrations involved in constructing a \app{} based on technologies like RAG and fine-tuning, particularly in specialized domains such as the legal field, healthcare, manufacturing, and others. 

These challenges span a wide range, from constructing data pipelines (\eg data processing and indexing) to leveraging LLMs' capabilities to achieve complex intelligent reasoning. For example, in applications of finance, there is a frequent need to understand and utilize high-dimensional time series data, whereas in healthcare, medical images or time-series medical records are often essential. Enabling LLMs to comprehend these varied forms of data represents a recurring challenge. On the other hand, in legal and mathematical applications, LLMs typically struggle to grasp long-distance dependencies between different structures. Additionally, depending on the specific application domain, there are increased demands for the interpretability and consistency of LLM responses. The inherent nature of LLMs tends to be characterized by low interpretability and high uncertainty, which poses significant challenges. Enhancing the transparency of LLMs and reducing their uncertainty are critical for increasing trust and reliability in their outputs, especially in fields where precision and accountability are paramount.

Through extensive discussions with domain experts and developers, and by carefully analyzing the challenges they face, we have gained a deep understanding that \app is not a one-size-fits-all solution. The real-world demands, particularly in expert domains, are highly complex and can vary significantly in their relationship with given data and the reasoning difficulties they require. However, developers often do not realize these distinctions and end up with a solution full of performance pitfalls (akin to a house with leaks everywhere). In contrast, if we could fully understand the demands at different levels and their unique challenges, we could build applications accordingly and make the application steadily improve (like constructing a solid and reliable house step by step).

\begin{figure}
    \centering
    \includegraphics[width=\textwidth]{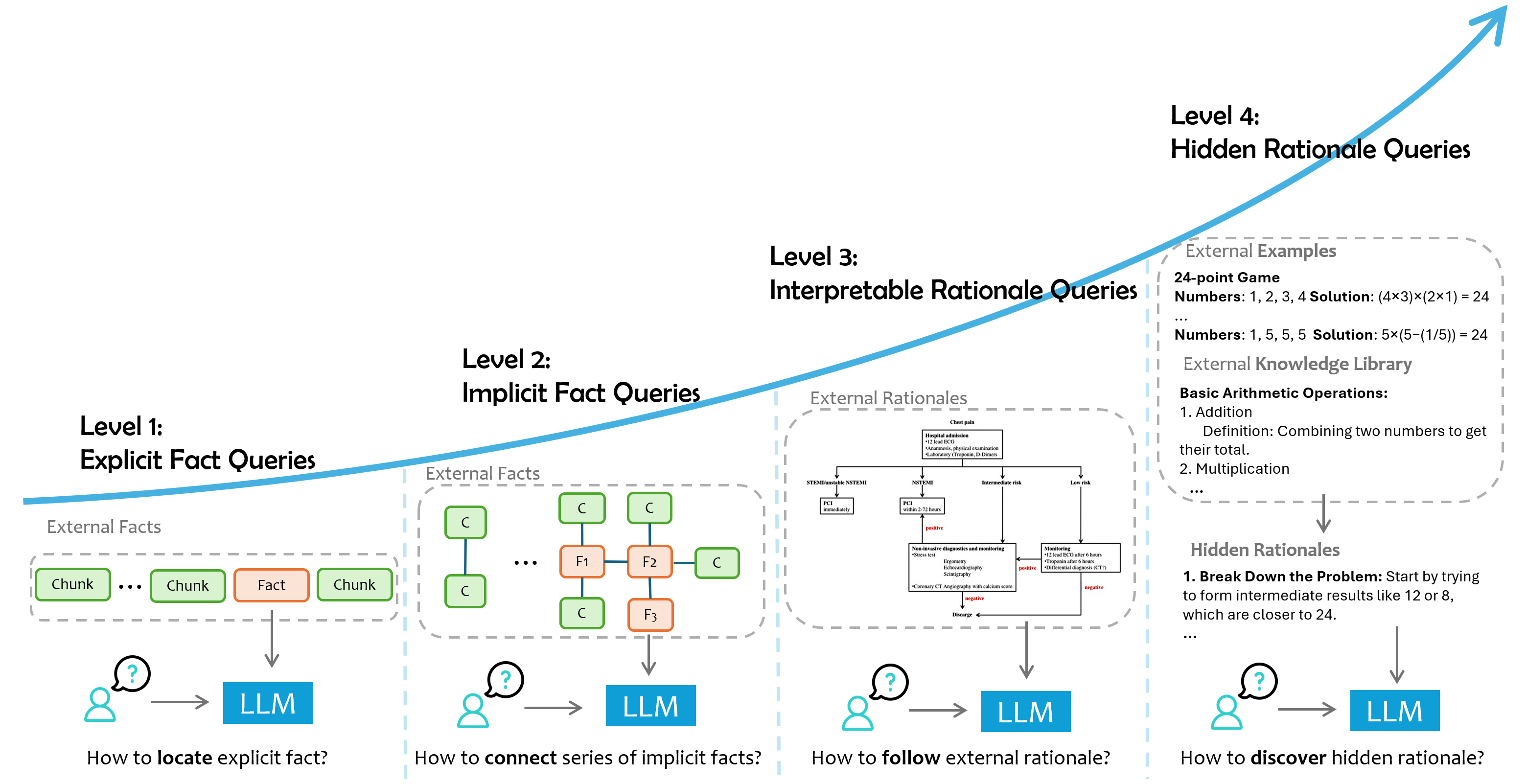}
    \caption{Main Focus of Four Level Queries}
    \label{fig:four-level-main-focus}
\end{figure}

Yet, research efforts and existing relevant surveys~\cite{rag_survey_23_1, rag_survey_24_2, rag_survey_24_3, rag_survey_24_4, rag_survey_24_5, finetune_survey_24_1, finetune_survey_23_2, finetune_survey_23_3} frequently focus on only one of these levels or a particular topic of technologies. This has motivated us to compile this comprehensive survey, which aims to clearly define these different levels of queries, identify the unique challenges associated with each(Figure~\ref{fig:four-level-main-focus}) , and list related works and efforts addressing them. This survey is intended to help readers construct a bird's-eye view of \app{} and also serve as a handbook on how to approach the development of such applications systematically. 

\section{Problem Definition}
Data-augmented LLM applications can take many forms, ranging from the frequently seen Question-Answering bots based on domain-specific data, to semantic processing operators within complex data pipelines, or even agents handling specific steps in a multi-agent system. However, in general, a data-augmented LLM application can be formulated as follows:
\begin{equation}
    f: \mathcal{Q} \xrightarrow{\mathcal{D}} \mathcal{A}
    \label{eq:da-app-def}
\end{equation}
where $\mathcal{Q}$, $\mathcal{A}$, and $\mathcal{D}$ represent the user's input (Query), the expected response (Answer), and the given data, respectively. The task of the application $f$ is to establish the mapping from $\mathcal{Q}$ to $\mathcal{A}$ based on $\mathcal{D}$.

In contrast to standalone LLM systems that rely solely on pre-existing knowledge, \app{} are characterized by their reliance on external data ($\mathcal{D}$) to accurately address the posed queries ($\mathcal{Q}$).
The incorporation of external data $\mathcal{D}$ can significantly bolster the capabilities of LLMs, granting them the ability to tap into current, domain-specific knowledge and to understand expert rationales. Queries can be stratified into various levels of complexity based on the extent and manner in which they utilize external data, reflecting the depth and nature of engagement required by the queries.

\subsection{Stratification of Queries}
In the landscape of data-augmented LLM applications, queries can be stratified based on their complexity and the depth of data interaction required. This stratification helps in understanding the varying levels of cognitive processing that an LLM must perform to generate accurate and relevant responses. From straightforward fact retrieval to the nuanced interpretation of implicit knowledge, each level represents a step up in the sophistication of the tasks that LLMs are expected to handle. Below, we delineate these levels, providing insights into the unique challenges and capabilities necessitated at each stage.

\begin{figure}
    \centering
    \includegraphics[width=\textwidth]{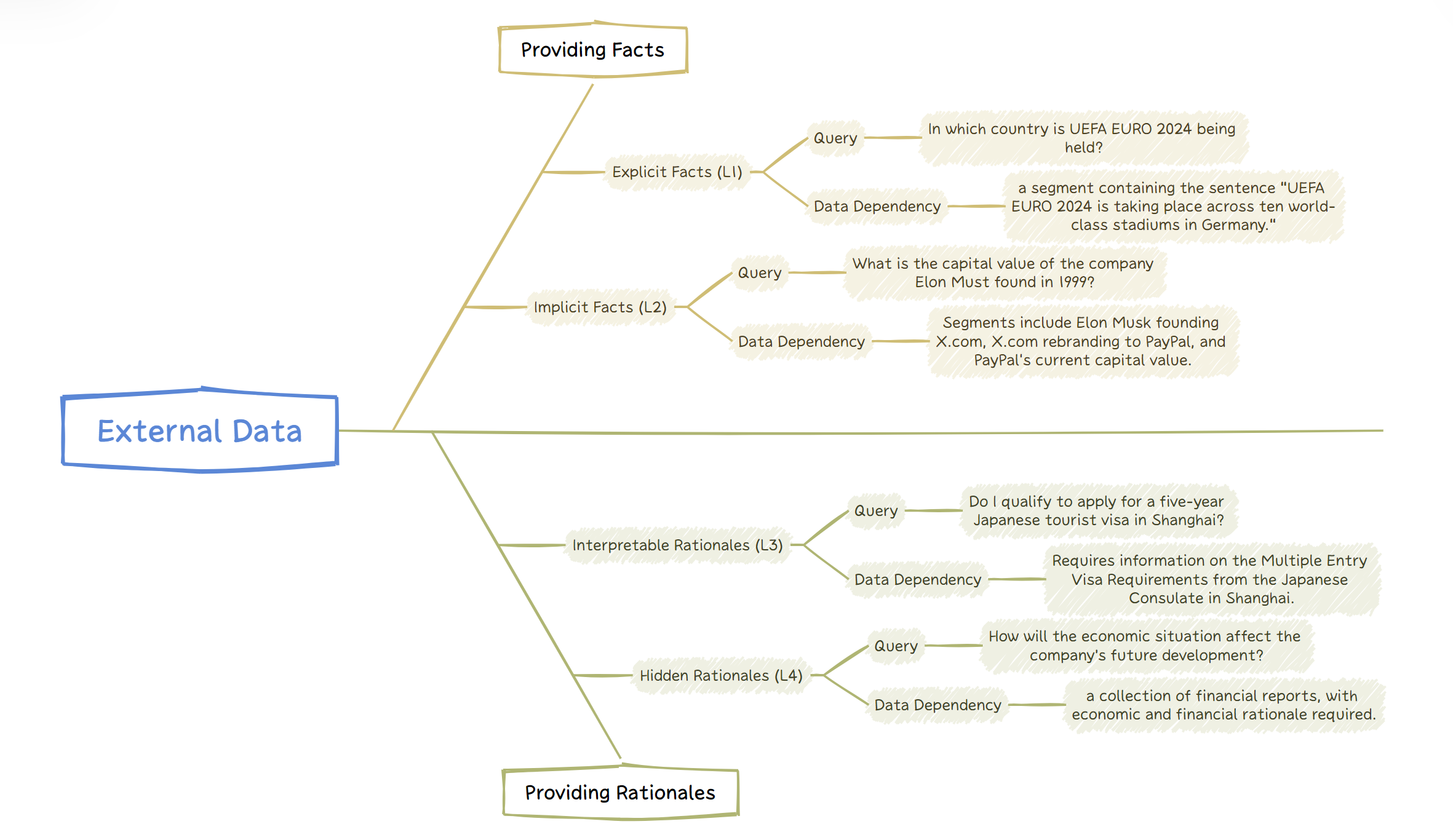}
    \caption{Summary of Query Levels in \MFUsentencecase{\app}}
    \label{fig:levels-examples}
\end{figure}

\begin{enumerate}[label=\textbf{Level-\arabic*}]
    
    \item \textbf{Explicit Facts}: These queries are asking about \textbf{explicit facts} directly present in the given data without requiring any additional reasoning. This is the simplest form of query, where the model's task is primarily to locate and extract the relevant information. For example, \textit{"Where will the 2024 Summer Olympics be held?"} targets a fact contained in the external data.
    
    \item \textbf{Implicit Facts}: These queries ask about \textbf{implicit facts} in the data, which are not immediately obvious and may require some level of common sense reasoning or basic logical deductions. The necessary information might be spread across multiple segments or require simple inferencing. For instance, the question \textit{"What is the majority party now in the country where Canberra is located?"} can be answered by combining the fact that Canberra is in Australia with the information about the current majority party in Australia.
    
    \item \textbf{Interpretable Rationales}: These queries demand not only a grasp of the factual content but also the capacity to comprehend and apply domain-specific \textbf{rationales} that are integral to the data's context. These rationales are often explicitly provided in external resources and is typically not present or rarely encountered during the pre-training phase of a general large language model. For example, in the realm of pharmaceuticals, an LLM must interpret FDA Guidance\footnote{\url{https://www.fda.gov/industry/fda-basics-industry/guidances}} documents—which represent the FDA's current thinking—to evaluate whether a specific drug application adheres to regulatory requirements. Similarly, in customer support scenarios, the LLM must navigate the intricacies of a predefined workflow to process user inquiries effectively. In the medical field, many diagnostic manuals provide authoritative and standardized diagnostic criteria, such as management guidelines for patients with acute chest pain~\cite{gruettner2012clinical}. By effectively following these external rationales, it is possible to develop a specialized LLM expert system for managing chest pain. This involves understanding the procedural steps and decision trees that guide a support agent's interactions with customers, ensuring responses are not only accurate but also comply with the company's service standards and protocols. 
    
    \item \textbf{Hidden Rationales}: This category of queries delves into the more challenging realm where the rationales are not explicitly documented but must be inferred from patterns and outcomes observed in external data. The hidden rationales here refer not only to the implicit reasoning chains and logical relationships, but also to the inherently challenging and non-trivial task of identifying and extracting the external rationales required for each specific query. In IT operational scenarios, for example, a cloud operations team may have addressed numerous incidents in the past, each with its own unique set of circumstances and resolutions. The LLM must be adept at mining this rich repository of tacit knowledge to discern the implicit strategies and decision-making processes that were successful. Similarly, in software development, the debugging history of previous bugs can provide a wealth of implicit insights. While the step-by-step rationale for each debugging decision may not be systematically recorded, the LLM must be capable of extracting the underlying principles that guided those decisions. By synthesizing these hidden rationales, the LLM can generate responses that are not only accurate but also reflective of the unspoken expertise and problem-solving approaches that have been honed over time by experienced professionals.

\end{enumerate}

In summary, the classification of queries into levels reflects a gradient of complexity and the type of understanding required from the LLM. As shown in Figure~\ref{fig:four-level-main-focus} and exampled by Figure~\ref{fig:levels-examples}, the first two levels, \textbf{Explicit Facts} and \textbf{Implicit Facts}, focus on the retrieval of factual information, whether directly stated or requiring basic inferencing. These levels challenge the LLM's ability to extract and synthesize data into coherent facts.
Conversely, the latter two levels, \textbf{Interpretable Rationales} and \textbf{Hidden Rationales}, shift the focus towards the LLM's capacity to learn and apply the rationales behind the data. These levels demand a deeper cognitive engagement, where the LLM must align with expert thinking or extract wisdom from unstructured historical data, respectively. The classification of common factual querying datasets according to this standard is depicted in Table ~\ref{tab: dataset_categorization}.

\begin{table}[]
\centering
\begin{tabular}{cccc}
\hline
Task   Categorization & \rule{0pt}{1.2em} Datasets   \vspace{0.2em}                              & levels             & Mutiple References \\ \hline
\multirow{17}{*}{QA} & \rule{0pt}{1em} NQ(Natural Questions)\cite{kwiatkowski2019natural}     & 1 - Explicit Facts & False \\
                      & \rule{0pt}{1em}MS MARCO \cite{bajaj2016ms}              & 1 - Explicit Facts & False              \\
                      & \rule{0pt}{1em}TriviaQA\cite{2017arXivtriviaqa}         & 1 - Explicit Facts & False              \\
                      & \rule{0pt}{1em}SQuAD~\cite{rajpurkar2016squad}          & 1 - Explicit Facts & False              \\
                      & \rule{0pt}{1em}ASQA~\cite{stelmakh2022asqa}             & 1 - Explicit Facts & False              \\
                      & \rule{0pt}{1em}WebQSP~\cite{webqsp_dataset}             & 1 - Explicit Facts & False              \\
                      & \rule{0pt}{1em}HotPotQA\cite{yang2018hotpotqa}          & 2 - Implicit Facts & True               \\
                      & \rule{0pt}{1em}2WikiMultiHopQA\cite{ho2020constructing} & 2 - Implicit Facts & True               \\
                      & \rule{0pt}{1em}MuSiQue\cite{trivedi2022musique}         & 2 - Implicit Facts & True               \\
                      & \rule{0pt}{1em}Bamboogle\cite{press2022measuring}       & 2 - Implicit Facts & True               \\
                      & \rule{0pt}{1em}StrategyQA\cite{geva2021did}             & 2 - Implicit Facts & True               \\
                      & \rule{0pt}{1em}ComplexWebQuestions~\cite{ComplexWebQuestions_dataset} & 2 - Implicit Facts & True  \\
                      & \rule{0pt}{1em}WebQuestions~\cite{webquestion_dataset}  & 2 - Implicit Facts & True               \\
                      & \rule{0pt}{1em}Mintaka~\cite{sen2022mintaka}            & 2 - Implicit Facts & True               \\
                      & \rule{0pt}{1em}MetaQA~\cite{metaqa_dataset}             & 2 - Implicit Facts & True               \\
                      & \rule{0pt}{1em}qasper~\cite{Dasigi2021ADO}              & 2 - Implicit Facts & True               \\
                      & \rule{0pt}{1em}DROP~\cite{dua2019drop}                  & 2 - Implicit Facts & True               \\
Multi-Choice          & \rule{0pt}{1em}QuALITY~\cite{pang2021quality}           & 2 - Implicit Facts & True               \\
Fact Checking         & \rule{0pt}{1em} Feverous\cite{aly2021feverous} \vspace{0.2em}          & 2 - Implicit Facts & True               \\ \hline
\end{tabular}
\caption{Stratification of Common Datasets Providing Facts}
\label{tab: dataset_categorization}
\end{table}

Each level presents its unique set of challenges and, consequently, necessitates tailored solutions to effectively address them. As we delve into the intricacies of these levels in the following sections, we will explore the specific strategies and methodologies that enable LLMs to navigate the complexities of data-augmented applications across these varied spectrums of query types. This exploration will not only highlight the current capabilities of LLMs but also shed light on the ongoing advancements and potential future developments in the field.

\section{Explicit Fact Queries (L1)}

\subsection{Overview}

\MFUsentencecase{\firstlevel}, represent the most straightforward type of data-augmented queries. Queries at this level can be answered by directly accessing specific domain documents or document snippets within the collection. The answers to these questions are often in plain text within the documents, requiring minimal reasoning or simple rationale in the response generation.

The defining characteristic of this level is the clear and direct dependency on specific pieces of external data. 

\subsubsection{Data Dependency}

The dataset $\mathcal{D}$ can be segmented into documents or segments, denoted as $D_1, D_2, \ldots, D_n$, in various ways:
\begin{equation}
    \mathcal{D} = \left\{ D_1, D_2, \ldots, D_n \right\}
\end{equation}
Each segment $D_i$ is considered relatively short and contains content that is more focused and specific\footnote{In some most recent advancements, the segment size may be as large as a whole document or even larger}. 

For a given query \(q \in \mathcal{Q}\), not every segment within $\mathcal{D}$ is requisite for formulating a response. Let $\delta: \mathcal{Q} \times \mathcal{D} \rightarrow \{0,1\}$ denote the necessity of data segment \(d \in \mathcal{D}\) for a specific query \(q\), where $\delta(q,d) = 1$ means that data segment \(d\) is required to answer the query \(q\), and $\delta(q,d) = 0$ otherwise. Then the data dependency of query \(q\), characterized by the subset of segments indispensable for addressing query \(q\), is defined as:
\begin{equation}
    Dep(q) = \{d \mid d \in \mathcal{D} \text{ and } \delta(q,d) = 1\} 
\end{equation}
It's easy to understand that $Dep(q) \in \mathcal{P}(\mathcal{D})$, where $\mathcal{P}(\mathcal{D})$ is the power set\footnote{The power set (or powerset) of a set $S$ is the set of all subsets of $S$, including the empty set and $S$ itself.} of $\mathcal{D}$.

\subsubsection{Definition}
\MFUsentencecase{\firstlevel}, denoted as \(\mathcal{Q}_1\), are characterized by the direct retrievability of answers from specific data segments within the dataset \(\mathcal{D}\). These queries can be formally defined in the context of a data-augmented LLM system as follows:

For any query \( q \) and its corresponding answer \( a \), an explicit fact query is one where there exists:
\begin{itemize}
    \item A retrieval component \( r_{\mathcal{D}}: \mathcal{Q} \rightarrow \mathcal{P}(\mathcal{D}) \) that identifies the relevant data segments from \(\mathcal{D}\) necessary to answer \( q \). This component ensures that \( r_{\mathcal{D}}(q) \) closely matches \( Dep(q) \), the minimal subset of \(\mathcal{D}\) required to respond to \( q \).
    \item A response generator \( \theta \), typically a prompted LLM inference, that constructs the answer \( a \) based solely on the information retrieved by \( r_{\mathcal{D}} \). The response \( \theta(r_{\mathcal{D}}(q)) \) should be equal to or approximate \( a \), demonstrating the query's reliance on explicit, directly accessible facts.
\end{itemize}
This definition underscores the reliance of \firstlevel on direct data retrieval without the need for complex reasoning or inference beyond the scope of the identified data segments.

Here are some examples of queries at this level:
\begin{itemize}
    \item \textit{What method was used in Paper X to solve problem Y?} (given a collection of academic papers)
    \item \textit{What's the AI strategy of company X?} (given a series of the latest news and articles about company X)
\end{itemize}

\subsection{Challenges and Solutions}
Queries at this level primarily necessitate the correct retrieval of data for LLMs to provide accurate responses. RAG~\cite{rag_survey_23_1}, due to its effectiveness, flexibility, and relatively low costs, is the most commonly adopted technical solution for handling this level of queries. However, even with RAG, there are significant challenges in constructing a robust and high-quality system. These challenges include:

\begin{itemize}
    \item \textbf{Data Processing Difficulties}: External data is often highly unstructured and contains multi-modal components such as tables, images, videos, and more. Additionally, the process of segmenting or "chunking" this data presents challenges in maintaining the original context and meaning.
    
    \item \textbf{Data Retrieval Difficulties}: The retrieval of relevant data segments from a large, unstructured dataset can be computationally intensive and prone to errors. The challenge lies in developing efficient and accurate retrieval mechanisms.
    
    \item \textbf{Evaluation Difficulties}: Evaluating the performance of a RAG system, particularly at a component level, is a complex task. It requires the development of robust metrics that can accurately assess the quality of data retrieval and response generation.
\end{itemize}

Given the popularity of RAG, a wealth of literature and tools have been developed to address these challenges. In the remainder of this section, we will highlight some of the most practical and impactful enhancements to RAG. Additionally, we will discuss alternative technical solutions that may be employed beyond RAG.

\subsection{Retrieval-augmented Generation (RAG)}

Retrieval-Augmented Generation refers to a methodology where a language model augments its natural language generation capabilities by dynamically retrieving external information during the generation process. This technique blends the generative capabilities of LLMs with the information retrieval from extensive databases or documents. The process is typically implemented as data index construction, retrieval system construction and answer generation.

\subsubsection{Data Processing Enhancement}

Document parsing at this level often involves extracting information from text, tables, and figures in a coherent manner, ensuring that the relevant snippets are accurately identified and retrieved.

\textit{\textbf{Multi-modal Documents Parsing}} Addressing multi-modal content in source documents, such as charts, tables, or even videos (e.g. meeting recordings), is one of the most frequently asked questions. Broadly, two approaches are employed to tackle this issue. The first approach involves converting multi-modal content into textual form. For instance, Table-to-Text methods~\cite{min2024exploring} translate tables into text, while other techniques convert visual content into textual or attribute-based descriptions~\cite{suris2023vipergpt, gao2023assistgpt}, which are subsequently processed by large language models. The second approach leverages multi-modal embedding techniques~\cite{hu2023reveal, li2022blip, long2024generative}, utilizing the retrieved embeddings from multi-modal data as soft prompts for input.

\textit{\textbf{Chunking Optimization}} For long texts, segmenting documents into text chunks is a common and necessary operation. Larger text chunks can preserve more of the semantic coherence of the context, but they also tend to contain more noise within each chunk\cite{liu2024lost}. Commonly-used chunking strategies~\cite{langchain_chunk, llamaindex_chunk} include fixed size chunking, recursive chunking, sliding window chunking, paragraph-based chunking ,semantic chunking, etc. Certain methods are designed to ascertain the level of detail a query demands and, based on this identification, select text chunks of appropriate granularity for retrieval\cite{sarthi2024raptor, zhong2024mix}. Alternatively, some methods opt to process and refine the text into smaller segments that maintain a high degree of information completeness\cite{chen2023dense}. Additionally, there are approaches that employ vision models to segment text in accordance with the original document structure\cite{yepes2024financial}.

\textit{\textbf{}} 
\subsubsection{Data Retrieval Enhancement}
Information Retrieval (IR) techniques can be smoothly transferred into RAG applicaitons. The primary steps involved include establishing data indexes, processing queries, retrieving and matching, re-ranking, and evaluation.

\textit{\textbf{Indexing}} The purpose of this step is to establish mappings from search terms to text segments, determining the logic by which the retrieval system operates. Indexing methods are broadly classified into three types: sparse, dense, and hybrid retrieval. Sparse retrieval uses specific words to index text segments. In contrast, dense retrieval maps text segments into a dense vector space of features. Hybrid retrieval combines elements of both sparse and dense techniques.

\begin{itemize}
    \item \textit{Sparse Retrieval}: This was the first indexing method to be widely adopted due to its simplicity and intuitiveness. Techniques like TF-IDF and BM25\cite{sparck1972statistical, robertson1995okapi} are designed to identify the most representative keywords of each text segment based on their relative frequency. These methods are still prevalent in many RAG projects\cite{ram2023context, jiang2023active, xu2024recomp}. However, word matching methods can lead to retrieval losses due to their inability to recognize synonyms. To address this issue, methods like KNN can be used for similarity-based matching of keywords \cite{boytsov2016off}. Alternatively, indices like keywords can be changed into the prediction of the probabilities of query tokens for the corresponding text segment\cite{zhuang2021tilde, zhuang2021fast}.
    \item  \textit{Dense Retrieval}: This approach often involves using pre-trained or fine-tuned text encoders to map texts to a dense vector space that aligns with query requirements. BERT-based encoders~\cite{devlin2018bert} are commonly to be fine-tuned as dense retriever on unsupervised data using methods such as DPR\cite{karpukhin2020dense}, ANCE\cite{xiong2020approximate}, SimCSE\cite{gao2021simcse} and TAS-B\cite{hofstatter2021efficiently}. Others employ unsupervised contrastive learning for fine-tuning, such as  Contriever\cite{izacard2021unsupervised}. Using feedback from LLMs to guide the training objectives of retrievers can also effectively enhance the retriever's suitability for LLMs \cite{zhang2023retrieve, ma2023fine, borgeaud2022improving}. Given the powerful capabilities and expressive potential of LLMs, LLM-based dense retrieval has recently emerged as a key area of interest and exploration~\cite{ni2021large}. LLM2vec~\cite{behnamghader2024llm2vec} modifies the attention mechanism of a pre-trained LLM to a bidirectional one and employs the masked next-token prediction method for unsupervised training, resulting in an LLM-based dense retrieval embedder. Similarly, Llama2Vec~\cite{li2024llama2vec} leverages two pretext tasks—Embedding-Based Auto-Encoding and Embedding-Based Auto-Regression—to train an unsupervised dense retrieval encoder based on the LLaMA architecture~\cite{touvron2023llama}, leading to significant improvements in retrieval task performance.
    \item \textit{Others}: Combining sparse retrieval and dense retrieval is an effective method to focus simultaneously on the central theme of text segments and global features. Feng et al. (2023) propose initially determining the knowledge domain needed to answer a query as a fixed area of expertise, and then using dense retrieval to recall supplementary information within this domain \cite{feng2023knowledge}. Numerous studies have explored various methods of blending dense vector indexing with sparse encoder indexing to better capture the semantic information of text blocks and enhance the precision of targeted paragraph retrieval \cite{sawarkar2024blended, chen2024bge, luan2021sparse}. On the other hand, Tang et al. (2024) have enhanced the capabilities of a LLM by fine-tuning it for indexing and retrieving, effectively integrating these abilities directly into the LLM. This allows the LLM to autonomously generate data indices and text segments for each query \cite{tang2024self, cheng2024lift}.
\end{itemize}

\begin{figure}
    \centering
    \includegraphics[width=\textwidth]{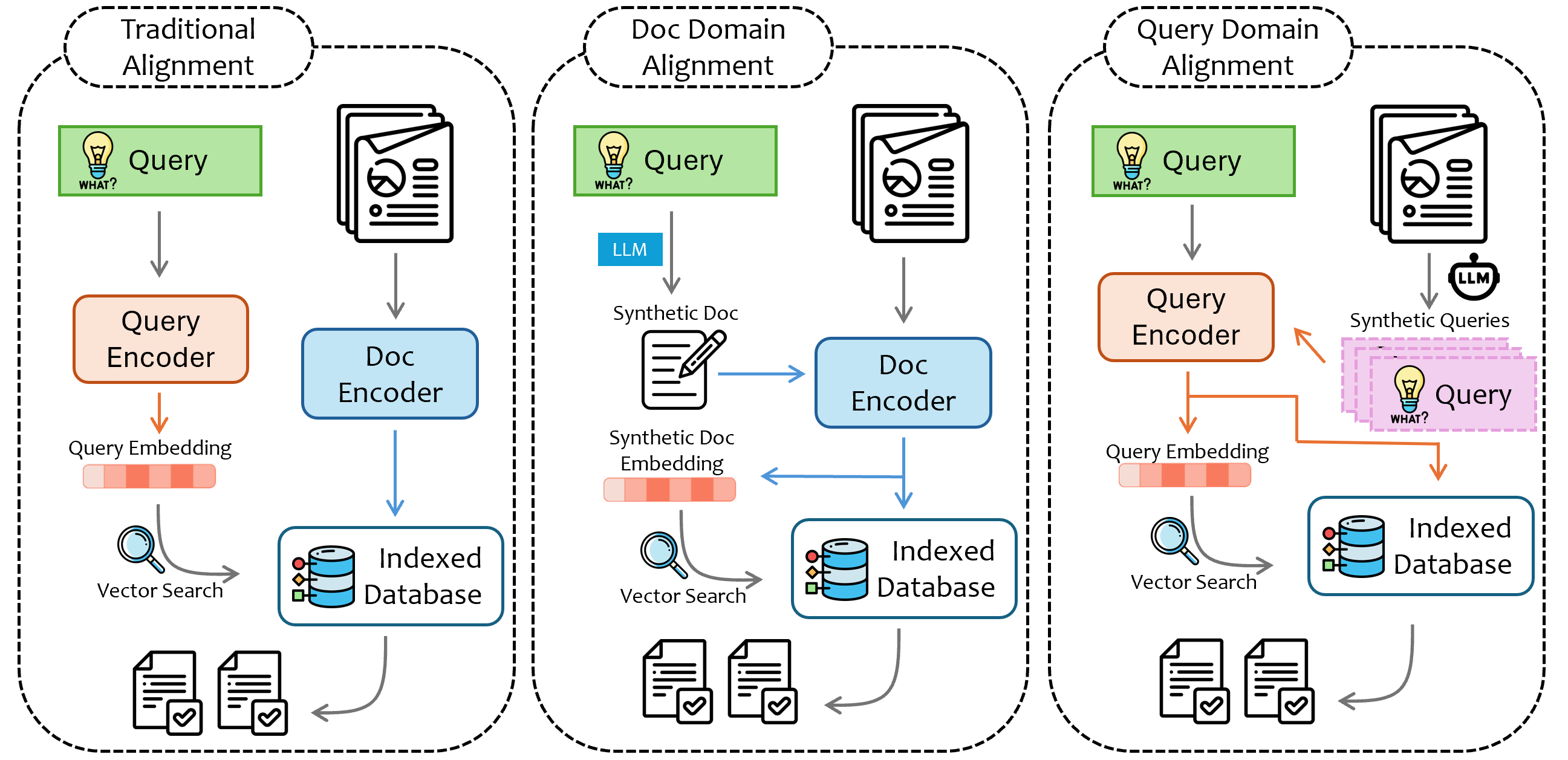}
    \caption{Three Types of Query-Document Alignment}
    \label{fig:three-alignments}
\end{figure}

\textit{\textbf{Query Document Alignment}}  The goal of this step is to align the query with document segments in external data to identify the best document segment that can assist in answering the query. As Figure~\ref{fig:three-alignments} illustrated, there are primarily three approaches to this alignment: traditional alignment, document domain alignment, and query domain alignment. Traditional alignment involves mapping both document segments and the query into the same encoding space. For instance, many dense retrieval architectures based on dual encoders feature specialized query encoders~\cite{karpukhin2020dense, xiong2020approximate, hofstatter2021efficiently}. Conversely, if a system like RAG employs sparse retrieval, it is necessary to extract keywords from the query for the search. Further refinement can be achieved through query rewriting techniques, which enhance search accuracy by mitigating issues related to user terminological inaccuracies or vague descriptions, effectively improving the precision of the search results~\cite{ma2023query}. Document domain alignment involves generating synthetic answers first, then using these answers to recall relevant data, effectively addressing the issue of queries and retrieved data not being in the same distribution space. A notable work in this area is HyDE~\cite{gao2022precise}. Query domain alignment~\cite{raina2024question} involves generating a set of synthetic questions for each atomic unit of text, mapping text segments into the query space, and then retrieving the synthetic questions closest to the original query along with their corresponding text segments. This method ensures that the most relevant and contextually appropriate segments are selected for responding to the query. SlimPLM~\cite{tan2024small} employs a small proxy model to generate heuristic answers, which are then used to predict the knowledge needed to answer the question. This approach also provides an effective method for aligning queries to the document space.

\textit{\textbf{Re-ranking and Correction}} 
After retrieving the top k text blocks, RAG systems must filter and reorder these segments. Most RAG systems use the relevance scores provided by the retriever as the basis for ranking, while some studies employ specific metrics such as perplexity or perplexity gain as ranking criteria \cite{jiang2023llmlingua, jiang2023longllmlingua}. Other efforts involve using LLMs to evaluate the credibility and utility of retrieved text blocks, training a pluggable reward-driven contextual adapter to refine the output of retriever\cite{yang2023prca}. Additionally, some research focuses on pre-training a small language model dedicated to fact verification, which is used to filter out incorrect retrieved text chunks, thus improving the quality of the recalled text\cite{yan2024corrective}.

\textit{\textbf{Recursive Retrieval or Iterative Retrieval}}
Considering the inherent limitations in the accuracy of a single retrieval attempt, an effective mitigation strategy is to perform multiple retrievals to progressively address any omissions. Kim et al. (2023) introduced a tree-like recursive retrieval method, incorporating pruning strategies to incrementally break down ambiguous questions into disambiguated ones, ultimately arriving at the closest correct answer \cite{kim2023tree}. Similarly, SEATER uses the k-means algorithm to construct a hierarchical tree structure of items to be retrieved, and iteratively recalls nodes within the tree structure \cite{si2023generative}.

\subsubsection{Response Generation Enhancement}
Generating responses requires determining if the retrieved information is sufficient or if additional external data is needed. Handling conflicts between retrieved knowledge and the model's internal prior knowledge is also essential \cite{wu2024faithful, ding2024retrieve, tan2024blinded}. Supervised fine-tuning is an effective method to enhance the generation performance in RAG systems. When faced with irrelevant or erroneous information as the retrieved context, pre-trained large language models are often easily misled, resulting in incorrect responses. Many studies have shown that by subtly designing training data for RAG systems, fine-tuning or pretraining can effectively mitigate this issue \cite{wang2023instructretro, yoran2023making, fang2024enhancing}. Through experimental analysis, RAAT~\cite{fang2024enhancing}, demonstrated that the detrimental effects of irrelevant retrieval noise, relevant retrieval noise, and counterfactual retrieval noise on RAG models increase progressively. By incorporating with these training process, these methods enables the LLM to internally recognize noisy contexts, leading to significant improvements in response generation quality even in the presence of noisy retrievals. Furthermore, to ensure more consistent performance between the retriever and generator within the RAG system, some studies employ joint training of both retriever and generator during the training phase \cite{lin2023ra, hofstatter2023fid, asai2023self}.

\section{Implicit Fact Queries (L2)}

\subsection{Overview}
These queries involve data dependencies that are not immediately obvious and may require some level of common sense reasoning or basic logical deductions. The necessary information might be spread across multiple segments or require simple inferencing.
(Example in Figure~\ref{fig:levels-examples})

Queries at this level require gathering and processing information from multiple documents within the collection. The collection of required information may exceed the ability of a single retrieval request, necessitating the decomposition of the original query into multiple retrieval operations and the aggregation of results into a comprehensive answer. This level often involves common-sense reasoning without requiring domain-specific expertise. This type of queries may include statistical queries, descriptive analysis queries, and basic aggregation queries. For example, operations such as counting, comparison, trend analysis, and selective summarization are common in "how many" and "what's the most" type queries, while multi-hop reasoning is frequently used. Therefore, we can define the level-2 queries, $\mathcal{Q}_2$ as follows:

For any query $q$ and its corresponding answer $a$, a $\mathcal{Q}_2$ fact query is one where:
\begin{itemize}
    \item There exists a set of \firstlevel \( \{q_1, q_2, \ldots, q_m\} \subset \mathcal{Q}_1 \), each of which can be directly retrieved from specific data segments within the dataset \( D \), such that:
    \[
    r_D(q) = \bigcup_{i=1}^{m} r_D(q_i)
    \]
    where \( r_D(q_i) \) identifies the relevant data segments from \( D \) necessary to answer \( q_i \), and the union of these segments provides the information necessary to answer \( q \).
    
    \item A response generator \( \theta \), typically a prompted LLM inference, constructs the answer \( a \) to \( q \) by aggregating the responses \( \{\theta(r_D(q_1)), \theta(r_D(q_2)), \ldots, \theta(r_D(q_m))\} \) and applying common-sense reasoning to derive an answer that is not explicitly stated in the data. The response \( \theta(r_D(q)) \) should approximate the correct answer \( a \), demonstrating that the query \( q \) can be effectively answered through the aggregation of responses to the \( \mathcal{Q}_1 \) queries.
\end{itemize}

This definition underscores the reliance of $\mathcal{Q}_2$ queries on the ability to decompose complex queries into a set of simpler, \firstlevel \( \mathcal{Q}_1 \), whose answers can then be combined to generate the correct response to the original query \( \mathcal{Q}_2 \).

Here are some examples of queries at this level:
\begin{itemize}
    \item \textit{How many experiments have sample sizes greater than 1000?} (given a collection of experimental records)
    \item \textit{What are the top 3 most frequently mentioned symptoms?} (given a collection of medical records)
    \item \textit{What's the difference between the AI strategies of company X and company Y?} (given a series of the latest news and articles about companies X and Y)
\end{itemize}

\subsection{Challenges and Solutions}
At this level, queries still revolve around factual questions, but the answers are not explicitly presented in any single text passage. Instead, they require combining multiple facts through common-sense reasoning to arrive at a conclusion. The challenges of a level-2 query primarily include:

\begin{itemize}
    \item \textbf{Adaptive retrieval volumes}: Different questions may require varying numbers of retrieved contexts, and the specific number of retrieved contexts can depend on both the question and the dataset. A fixed number of retrievals may result in either information noise or insufficient information.
    \item  \textbf{Coordination between reasoning and retrieval}: Reasoning can guide the focus of what needs to be retrieved,  while the insights gained from retrieved information can iteratively refine reasoning strategies. Addressing these complexities calls for an intelligent integration and selective harnessing of external data, capitalizing on the inherent reasoning prowess of LLMs.
\end{itemize}

Methods to address challenges at this level include iterative RAG, RAG on graph/tree, and RAG with SQL, among others.

\subsection{Iterative RAG}


\MFUsentencecase{\secondlevel} is similar to multi-hop RAG tasks. This category of methods dynamically controls multi-step RAG processes, iteratively gathering or correcting information until the correct answer is achieved.

\begin{itemize}
    \item \textit{Planning-based}: Generating a stepwise retrieval plan during the prior-retrieval stage or dynamically within the retrieval process can refine the focus of each retrieval, efficiently guiding the iterative RAG system. For example, ReAct \cite{yao2022react} progressively updates the target of each step, reducing the knowledge gap required to answer the question. IRCoT \cite{trivedi-etal-2023-ircot} and RAT\cite{wang2024rat} uses a Chain of Thought to guide the RAG pipeline, making decisions about the current retrieval target based on previously recalled information. GenGround~\cite{shi2024generate} enables LLMs to alternate between two stages until arriving at the final answer: (1) generating a simpler single-step question and producing a direct answer, and (2) tracing the question-answer pair back to the retrieved documents to verify and correct any inaccuracies in the predictions. This iterative process ensures more reliable and accurate responses.
    \item \textit{Information Gap Filling Based}: ITRG \cite{feng2024retrieval} introduces an iterative retrieval-generation collaboration framework, generating answers based on existing knowledge and then continuing to retrieve and generate for the unknown parts of the response in subsequent rounds. Similarly, FLARE \cite{jiang2023active} revisits and modifies low-probability tokens in answers generated in each iteration. On the other hand, Self-RAG \cite{asai2023self} fine-tunes a large model to autonomously decide when to search and when to stop searching and start answering questions.
\end{itemize}

\subsection{Graph/ Tree Question Answering}

Addressing \secondlevel requires synthesizing information from multiple references. Graphs or trees, whether knowledge-based or data-structured, naturally express the relational structure among texts, making them highly suitable for this type of data retrieval problem.

\begin{itemize}
    \item \textbf{Traditional Knowledge Graph}: One of the initial structures considered for enhancing the efficacy of LLMs is the traditional knowledge graph, where each node represents an entity and edges between nodes signify the relationships between these entities. 
    
    \cite{Pan2023UnifyingLL} proposed a forward-looking development roadmap for LLMs and Knowledge Graphs (KGs) comprising: 1) KG-enhanced LLMs, which integrate KGs during the pre-training and inference phases of LLMs to deepen the models' understanding of acquired knowledge; 2) LLM-enhanced KGs, which employ LLMs for various KG tasks such as embedding, completion, construction, graph-to-text generation, and question answering; and 3) collaborative LLMs+KGs approaches, where both LLMs and KGs play complementary roles, enhancing each other through bidirectional inference driven by data and knowledge. The Rigel-KQGA model~\cite{sen-etal-2023-knowledge}, is an end-to-end KGQA model that predicts the necessary knowledge graph nodes based on a query and combines this with an LLM to derive answers. Works like Think-on-Graph~\cite{sun2023thinkongraph}  and KnowledgeNavigator~\cite{guo2023knowledgenavigator} extract entities involved in a query and then perform iterative BFS searches on the graph, using the LLM as a thinking machine to determine the optimal exploration path and perform pruning. The $R^3$~\cite{r3}introduces several possible commonsense axioms via an LLM that could address a query, sequentially searching related knowledge subgraphs to assess if the current information suffices to answer the query, continuing until the question is resolved.
    
    \item \textbf{Data Chunk Graph/ Tree}: The impressive reading comprehension capabilities of LLMs enable them to effectively grasp text without needing to break it down into the finest granularities of entities and relationships. In this context, researchers have begun experimenting with using text chunks or data chunks as nodes on graphs or trees, employing edges to represent either high-level or more intricately designed relations. Knowledge-Graph-Prompting~\cite{23_kg_prompting} discusses three popular kinds of questions that require mining implicit facts from (a) bridging questions rely on sequential reasoning while (b) comparing questions rely on parallel reasoning over different passages. (c) structural questions rely on fetching contents in the corresponding document structures. To tackle these questions, Knowledge-Graph-Prompting utilizes entity recognition, TF-IDF, KNN, and document structure hierarchies to construct document graphs and extract subgraphs for answering questions. MoGG~\cite{zhong2024mix} treats one or two sentences as the smallest semantic units, using these as nodes and building edges based on semantic similarity between nodes. It also trains a predictor to determine the textual granularity required for answering a query by deciding how large a sub-graph is needed. To capture more high-level semantic relationships between text blocks, RAPTOR~\cite{sarthi2024raptor}, employs clustering algorithms to hierarchically cluster the finest granularity of text blocks. It summarizes new semantic information at each hierarchical level, recalling the most necessary information within a collapsed tree of nodes. Similarly, GraphRAG~\cite{24_graph_rag}, adopts a clustering approach. It initially connects the smallest text blocks based on semantic similarity, then uses community detection algorithms to group nodes. Finally, it summarizes the global answer to a query by analyzing responses within each node community.
\end{itemize}

\subsection{Natural Language to SQL Queries}
When dealing with structured data, converting natural language queries to SQL (NL2SQL) can be an effective approach. Tools like Chat2DB facilitate this process by translating user queries into database queries. In the era of large language models, there has been significant progress in the area of text-to-SQL\cite{shi2022xricl, li2023resdsql, poesia2022synchromesh, lin2020bridging}, which allows us to utilize these tools to retrieve information from structured databases. This capability serves as a valuable external data source to augment the generation capabilities of LLMs. By integrating text-to-SQL tools~\cite{biswal2024text2sql}, LLMs can access and incorporate structured data, enhancing their ability to generate more accurate and contextually relevant responses. This integration not only improves the depth and quality of the generated content but also expands the scope of LLM applications, enabling them to perform more complex tasks that require interaction with and interpretation of database content.

\subsection{Discussion on Fact Queries}
\textit{Whether to Use Fine-tuning}. Some works~\cite{fine_tune_new_24} have demonstrated the hardness of LLMs to acquire new factual knowledge during fine tuning. This process can lead to a deterioration in the overall performance of the LLMs in generating accurate responses, and it often results in the generation of more hallucinations. Furthermore, study~\cite{berglund2023reversal} suggests that fine-tuning LLMs with new factual data may cause the models to mechanically memorize fact statements. Interestingly, altering the phrasing of these memorized facts can render the recently learned knowledge ineffective, indicating a superficial level of understanding and retention by the LLMs. This points to limitations in the current fine-tuning processes and the need for more sophisticated methods to integrate and adapt new information effectively.

\textit{Whether to Separate Different Levels of Fact Queries}.  Both \firstlevel and \secondlevel are fact-based, and it is crucial to determine which level these queries belong before constructing \app. Misclassifying \firstlevel as \secondlevel can lead to the retrieval of an abundance of superficial information that is seemingly relevant but ultimately unhelpful for answering the question, which can mislead the LLM and waste computational resources. Conversely, mistaking \secondlevel for \firstlevel can prevent the use of appropriate methods to retrieve a sufficient and comprehensive set of external auxiliary data. \MFUsentencecase{\secondlevel} often require dynamically integrating information specific to the context of the queries, whereas \firstlevel generally need only a single data snippet, leading to the retrieval of a fixed amount of external data. This can result in suboptimal performance of the LLM. Therefore, it is advantageous to preliminarily distinguish the level of queries based on a thorough understanding of the target task. Additionally, considerable effort has been directed towards training models to autonomously assess whether the information retrieved is sufficient, exemplified by approaches such as self-RAG~\cite{asai2023self}.
\section{Interpretable Rationale Queries (L3)}

\subsection{Overview}
\begin{figure}[t]
    \centering
    \includegraphics[width=0.8\textwidth]{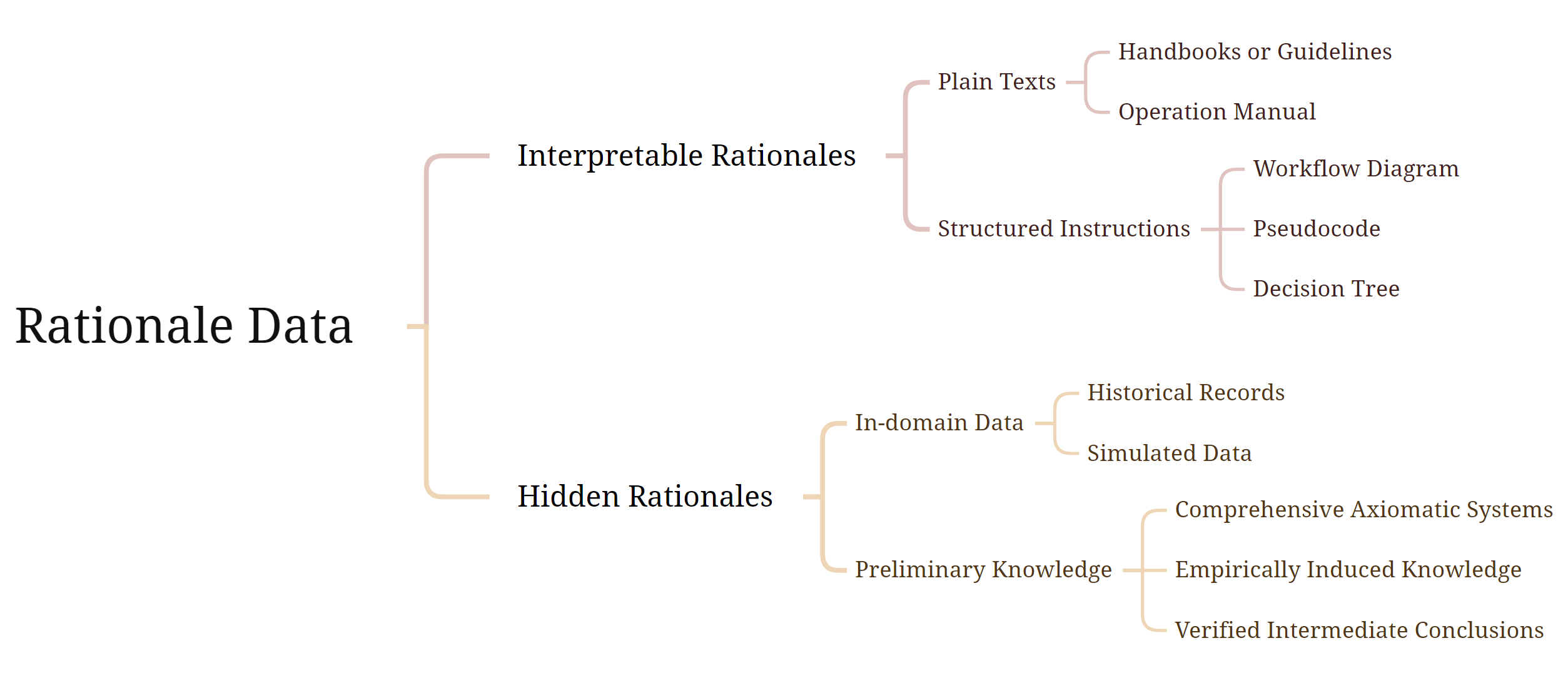}
    \caption{Demonstration of Rationale Queries}
    \label{fig:rationale}
\end{figure}

In this section and the next, we will explore queries that necessitate external data to furnish rationales for their resolution. These queries demand not only a grasp of the factual content but also the ability to comprehend and apply domain-specific \textbf{rationales} that are integral to the data's context. We classify these queries into two categories based on the nature of the rationales involved: queries based on interpretable rationales and those based on hidden rationales, as illustrated in the Figure ~\ref{fig:rationale}.

\MFUsentencecase{\thirdlevel} represent a relatively straightforward category within applications that rely on external data to provide rationales. The auxiliary data for these types of queries often include clear explanations of the thought processes used to solve problems. The data can be organized in several forms:

\begin{itemize}
    \item \textbf{Plain Texts}: Textual descriptions are the most common form of presenting interpretable rationales. These may include specialized or official documents such as handbooks or guidelines, as well as domain-specific manuals or operational guides. These texts articulate the reasoning processes that facilitate decision-making in complex scenarios. For example, documents such as FDA Guidance for pharmaceutical factories or medication guides for physicians provide insights into how experts, like FDA officers or doctors, approach specific cases.
    \item \textbf{Structured Instructions}: More explicit reasoning relationships or decision pathways might be presented in a structured format. These rationales can be understood as either a Text-Conditioned Moore Machine or a Text-Conditioned Mealy Machine. In the theory of computation, a Moore machine is a finite-state machine where the output values are determined solely by its current state\footnote{\url{https://en.wikipedia.org/wiki/Moore_machine}}. The conditions that control state transitions are often expressed in text, which LLMs need to interpret, unlike traditional programs that operate on native code. For instance, consider a customer supporting agent that follows a handbook to handle user's request to product changing or refunding. Similarly, a Mealy machine is a finite-state machine where output values are determined by both its current state and the inputs\footnote{\url{https://en.wikipedia.org/wiki/Mealy_machine}}. The distinction here is that actions (such as API calls) are determined not only by the state but also by the textual messages associated with transitions from the previous state. 
    Naturally, these domain-specific rationales can be represented in formats such as workflows, decision trees, or pseudocode.
\end{itemize}

Here are some examples of queries at this level:
\begin{itemize}
    \item \textit{How should a patient with chest pain and specific symptom descriptions be diagnosed and treated} (given a chest pain management guideline)
    \item \textit{How to respond to a user's question in a real-life scenario?} (given a customer service workflow)
\end{itemize}

\subsection{Challenges and Solutions}

In the realm of \thirdlevel, an additional challenge is integrating domain-specific rationales into LLMs in an comprehensible manner. The primary challenges are as follows:

\begin{itemize}
    \item \textbf{Prompt Optimization Costs}: The process of optimizing prompts is marked by high time and computational demands. Distinct queries demand tailored background knowledge and decision-making criteria, necessitating diverse examples. While manually designed prompts can be highly effective, they are labor-intensive and time-consuming. Furthermore, training models to generate tailored prompts for various queries incurs significant computational overhead.
    \item  \textbf{Limited interpretability}: The impact of prompts on LLMs is opaque. In many cases, access to the internal parameters of LLMs is typically restricted, complicating efforts to determine the impact of various prompts on these models. This lack of transparency hinders our ability to consistently understand and verify the interpretability of LLM responses to different prompts.
\end{itemize}

\subsection{Prompt Tuning}

For \thirdlevel, the key issue is how to effectively integrate rationales provided by external data into LLMs and ensure that these models can accurately follow and react based on these rationales. Text2MDT~\cite{zhu2024text2mdt} offers a viable demonstration, introducing two methods for automatically extracting medical decision trees from medical guidelines and textbooks. This process clarifies the logical chains within lengthy medical texts, making them more comprehensible. Similarly, MedDM~\cite{li2023meddm} has developed a format for clinical guidance trees that can be executed by LLMs, proposing a methodology for reasoning on these executable CGTs and a framework for multi-turn dialogues between patients and LLMs. InstructRec~\cite{zhang2023recommendation} aims to leverage the capabilities of LLMs in recommendation systems, designing a universal format to describe a user's preferences, intentions, task forms, and context using natural language, thereby creating a high-performing, language-based recommendation system.

Integrating rationales directly as natural language instructions into LLMs does not necessarily yield optimal performance, and manually designing prompts can be time-consuming. To address this, the employment of prompt tuning techniques becomes essential to enhance the LLMs' capability to adhere to specific rationales. One effective methodology is the application of reinforcement learning, as evidenced by the TEMPERA framework~\cite{zhang2022tempera}, which designs prompts incorporating limited instructions, examples, and verbalizers within the action space of reinforcement learning. Here, the LLM's probability of generating correct responses serves as the reward, guiding the model to discover the optimal prompt configuration across datasets. Similarly, Rlprompt~\cite{deng2022rlprompt} adopts a reinforcement learning approach, training an adaptor to assist smaller language models in producing optimal prompts based on feedback concerning the relative accuracy of LLM responses. Another innovative strategy, Directional Stimulus Prompting, leverages the performance of LLMs on downstream tasks as a reward mechanism. This method trains models to extract and utilize directional stimuli—specific cues or keywords tailored to individual instances—as prompts, thereby ensuring the LLMs' actions align more closely with expected outcomes.

Additionally, for optimization within discrete prompt spaces, edit-based methodologies such as GrIPS~\cite{prasad2022grips} are utilized. This technique involves using a small dataset as a scoring set to experiment with various prompt modifications—including deletions, swaps, paraphrases, and additions—to ascertain the most effective prompt configurations swiftly and effectively.

Recent advancements~\cite{zhou2022large, pryzant2023automatic} have also seen the rise of using LLMs themselves to facilitate prompt optimization. OPRO~\cite{yang2024large_1} employs an LLM both to generate new prompt solutions based on historical data and their associated performance metrics and to score these prompts, thus streamlining the optimization process. Furthermore, the Reflexion framework~\cite{shinn2024reflexion} introduces a novel prompt optimization approach based on linguistic feedback, using a language model to analyze and store reflections on LLM outputs in an episodic memory buffer. This memory component aids in refining decision-making processes and evaluating outcomes in future interactions, leveraging accumulated historical insights.

\subsection{CoT Prompting}

Addressing complex rationales necessitates that LLMs engage in extended chains of reasoning, a process distinct from the reasoning across disparate factual information typical of fact queries. However, Chain-of-Thoughts~\cite{wei2022chain}, Tree-of-Thoughts~\cite{yao2024tree} or Graph-of-Thoughts~\cite{yao2023beyond} methodologies proves effective for such scenarios. For issues that are well-studied and have high general applicability, manually designing CoT prompts emerges as a feasible solution. Ji et al. (2023)~\cite{ji2023towards} proposed a method of self-reflection that integrates knowledge acquisition with answer generation. By utilizing external tools and designing prompts, they constructed three types of self-reflection loops: the Factual Knowledge Acquiring Loop, the Knowledge-Consistent Answering Loop, and the Question-Entailment Answering Loop, thereby incorporating external rationales into the model's processing. Furthermore, Wu et al. (2024)~\cite{wu2024chain} conducted a manual analysis of error types in clinical records and developed three distinct CoT prompts to direct the GPT-4 model~\cite{achiam2023gpt} in focusing on intervention, diagnostic, and management errors. This targeted prompting facilitates the tasks of automatic error detection, span identification, and correction within clinical records.

While manual design of CoT prompts is highly effective, it requires substantial human and temporal resources. To mitigate these costs, Automate-CoT~\cite{shum2023automatic} proposed a technique for generating augmenting rational chains from a minimally labeled dataset. This approach employs a variance-reduced policy gradient strategy to evaluate the importance of each CoT chain, thus facilitating the selection of the most effective prompt combination.

Another form of utilizing Chain of Thoughts prompting involves constructing an agent workflow centered around LLMs. This typically requires the development of a more comprehensive system to address various real-world scenarios. According to Wang et al., such systems can be broadly divided into profiling, memory, planning, and action modules~\cite{wang2024survey}. Interpretable rationales can be integrated into multiple modules in various forms, allowing the agent to adapt and iterate based on environmental or human feedback. Recent advancements, such as those by LLM Reasoners~\cite{hao2024llm} and SocREval~\cite{he2024socreval}, have focused on automatically evaluating the quality of reasoning chains. These methodologies also assist in constructing robust \app{}.

Applications based on interpretable rationales span various domains. For instance, CoML~\cite{zhang-etal-2024-mlcopilot} integrates AutoML knowledge as prompts into an LLM, dynamically retrieves useful information from historical experimental records, and combines these elements to empower the LLM to develop machine learning solutions for novel tasks. MetaGPT~\cite{hong2023metagpt} has developed a multi-agent system for software development, where different stakeholders within a project are each represented as an agent. This setup enables multiple agents to collaborate according to a real-world work pipeline, effectively completing software development tasks. Similarly, sophisticated agent systems have been designed in fields such as customer service~\cite{abbasian2023conversational} and medical question answering~\cite{tang2023medagents}. In these domains, agents are tailored to handle specific types of inquiries, which can involve understanding complex user requests or providing accurate medical information. These systems not only enhance the interaction quality but also improve the efficiency and accuracy of responses, demonstrating the versatility and potential of LLMs when integrated into well-designed agent workflows.
\section{Hidden Rationale Queries (L4)}

\subsection{Overview}
\MFUsentencecase{\forthlevel} are the most challenging type of queries to address. Unlike \thirdlevel , which provide clear guidance on the rationales needed to respond to queries, \forthlevel involve domain-specific reasoning method that may not be explicitly described and are too numerous to exhaust. These rationales often encompass a wide variety that cannot be fully explored within the typical context window and may lack clear instructions, representing a form of domain expertise that is implicit within the data. Such data might include, but is not limited to:

\begin{itemize}
    \item \textbf{In-domain Data}: \MFUsentencecase{\forthlevel} might utilize data from the same domain, such as historical question-and-answer records or artificially generated data. This in-domain data inherently contains the reasoning skills or methodologies necessary to address current queries. For instance, in the context of Python programming puzzles, solutions to historical problems often include classical algorithms and problem-solving strategies that could aid in resolving current issues.
    \item \textbf{Preliminary Knowledge}: Another form of hidden rationales consists of extensive, dispersed knowledge bases that vary in application across different scenarios. This preliminary knowledge may constitute a comprehensive axiomatic system, such as all local legal codes that form the basis for legal judgments. It could also include proven intermediate conclusions that simplify reasoning processes in fields like mathematical proofs. When addressing real-world issues using external data, this prior knowledge might also stem from complex accumulations of human experiences and empirical summaries.
\end{itemize}

Navigating \forthlevel thus demands sophisticated analytical techniques to decode and leverage the latent wisdom embedded within disparate data sources, presenting significant challenges for RAG systems in effectively interpreting and applying such intricate and implicit information.

Here are some examples of queries at this level:
\begin{itemize}
    \item \textit{How will the economic situation affect the company's future development?} (given a collection of financial reports, with economic and financial rationale required)
    \item \textit{How to achieve 24 points using the numbers 5, 5, 5, and 1?} (given a series of 24-point game examples and corresponding answers.)
    \item \textit{Does Afghanistan permit a parent to confer his or her citizenship on a child born abroad?} (given the GLOBALCIT citizenship law dataset~\cite{guha2024legalbench})
\end{itemize}

\subsection{Challenges and Solutions}

The construction of data-augmented LLM applications is significantly challenged by \forthlevel, with primary difficulties manifesting in the following areas:

\begin{itemize}
    \item \textbf{Logical retrieval}: For questions involving hidden rationales, the helpfulness of external data does not simply depend on entity-level or semantic similarity, but rather on logical congruence or thematic alignment. Standard retrieval methods often struggle to capture the true target of the query or to identify text segments with logical similarities based on the problem presented. This necessitates the development of more sophisticated retrieval algorithms that can parse and identify underlying logical structures rather than relying solely on superficial textual similarities.
    \item  \textbf{Data insufficiency}: Fundamentally, external data may not explicitly contain the guidance or answers relevant to the current query. Instead, relevant information is often embedded in dispersed knowledges or illustrated through examples. This indirect presentation demands robust capabilities in data interpretation and synthesis, requiring LLMs to effectively derive coherent answers from fragmented or tangentially related data sources. Such challenges underscore the imperative for sophisticated data integration and reasoning capabilities within LLM frameworks to navigate the complexities of \forthlevel effectively.
\end{itemize}

\subsection{Offline Learning}
To address these types of queries, a common approach is to identify and extract rules and guidelines from datasets offline, subsequently retrieving related items. For generate reasoning rationales, some work like STaR~\cite{zelikman2022star} and LXS~\cite{stammer2023learning} used LLM for rationale generation. The former employs an iterative few-shot example method to generate from small dataset to large dataset, the later introduced a two-role explaining extraction process where a learner model generated explanations and a critic model assess them for validation. 

GL~\cite{pang2023guideline} identifies errors and generalizes them into guidelines for future tasks through an in-context-learning. LEAP~\cite{zhang2024context} forms principles by generating mistakes, low-level principles, and high-level principles, incorporating these principles into prompts for final inference. RICP~\cite{sun2024retrieved} uses mistakes from training data to generate high-level reasoning and specific insights, then employs hierarchical clustering to group modes of errors, generating task-level and question-level principles, which are combined and retrieved for question-level insights. A Buffer-of-Thought~\cite{yang2024buffer} uses a problem distiller to distill a meta-buffer across many reasoning tasks.

Some integrated methods, like MedPrompt~\cite{nori2023can}, include GPT-4-generated chains of thought for training examples with self-validation, using these in conjunction with a KNN retrieval in-context learning approach. Agent Hospital~\cite{li2024agent} generates rationales through reflection and utilizes both record retrieval and experience retrieval on generated data.

Although these concepts go by many different names—such as guidelines, principles, experiences, and thought templates—the main idea is to extract common useful rationales to enhance reasoning queries. These rationales may come from self-generated chains of thought (MedPrompt, Buffer-of-Thought), training set mistakes (GL, RICP, Agent Hospital), or intentionally generated mistakes (LEAP). Additionally, some principles are used across all tasks (Agent Hospital, RICP), while others are dynamically retrieved for specific questions (MedPrompt, Buffer-of-Thought). Many of these works demonstrate that learning from cases to accumulate experience as rationales is beneficial for various reasoning tasks.

\subsection{In Context Learning (ICL)}

Using examples for in-context learning is a common method for uncovering hidden rationales. Pre-trained large language models exhibit substantial in-context learning capabilities, which can be enhanced by retrieving examples based on similarity, thereby leveraging the models' few-shot learning abilities~\cite{liu2021makes, brown2020language}. However, the inclusion of irrelevant information in prompts can easily distract LLMs, leading to incorrect responses~\cite{shi2023large,chen2023many}. OpenICL, as developed by Wu et al.~\cite{wu2023openicl}, constructed an ICL framework that explores the impact of different traditional methods of retrieving examples and inference techniques on ICL effectiveness.

Furthermore, training smaller models based on the feedback from LLMs regarding context examples to select optimal demonstrations and examples can improve the construction of context for specific tasks in a more targeted manner~\cite{li2023unified, wang2024large, ye2023compositional}. To address the issue that semantic similarity-based example retrieval may not cover the broader range of associations needed in practical tests, Su et al.\cite{su2022selective} employed an unsupervised, graph-based selective annotation method called vote-k, constructing a more diverse and representative example database for few-shot learning. Additionally, Zhang et al.\cite{zhang2022automatic} proposed an Auto-CoT method that clusters examples into various representative types. By sampling problems diversely and generating reasoning chains, this method constructs examples that better support the learning process.

However, enabling LLMs to master reasoning capabilities outside their trained domains through few-shot learning remains a substantial challenge. Wang et al. addressed this by sampling a variety of reasoning paths and marginalizing over these paths to select the most consistent answer, thereby enhancing the probability of LLMs selecting correct reasoning chains~\cite{wang2022self}. Agarwal et al. introduced two scalable methods for generating usable example, namely reinforced ICL and unsupervised ICL, that aim to replace human-generated examples, thus expanding the pool of available examples~\cite{agarwal2024many}. DIN-SQL~\cite{pourreza2024din} sought to decompose tasks into simpler subtasks and used the solutions to these sub-problems as prompts for LLMs, significantly improving their performance in generating SQL from text. Similarly, DUP~\cite{zhong2024achieving} identified three main issues LLMs face when using the chain of thought approach to solve complex mathematical word problems: semantic misunderstandings, computational errors, and missing steps, with semantic misunderstandings being a primary limiting factor. Encouraging LLMs to deeply understand the problems and extract essential information for resolution can significantly enhance their ability to solve mathematical problems by addressing these semantic misunderstandings.

In-context learning is increasingly being utilized across various fields such as mathematics, law, medicine, and finance~\cite{schumacher2024team, dou2023plugmed, li2023large}, playing a crucial role in the development of data-augmented LLM applications. This approach not only extends the functional capabilities of LLMs but also enhances their practical utility across diverse domains.

\begin{figure}
    \centering
    \includegraphics[width=\textwidth]{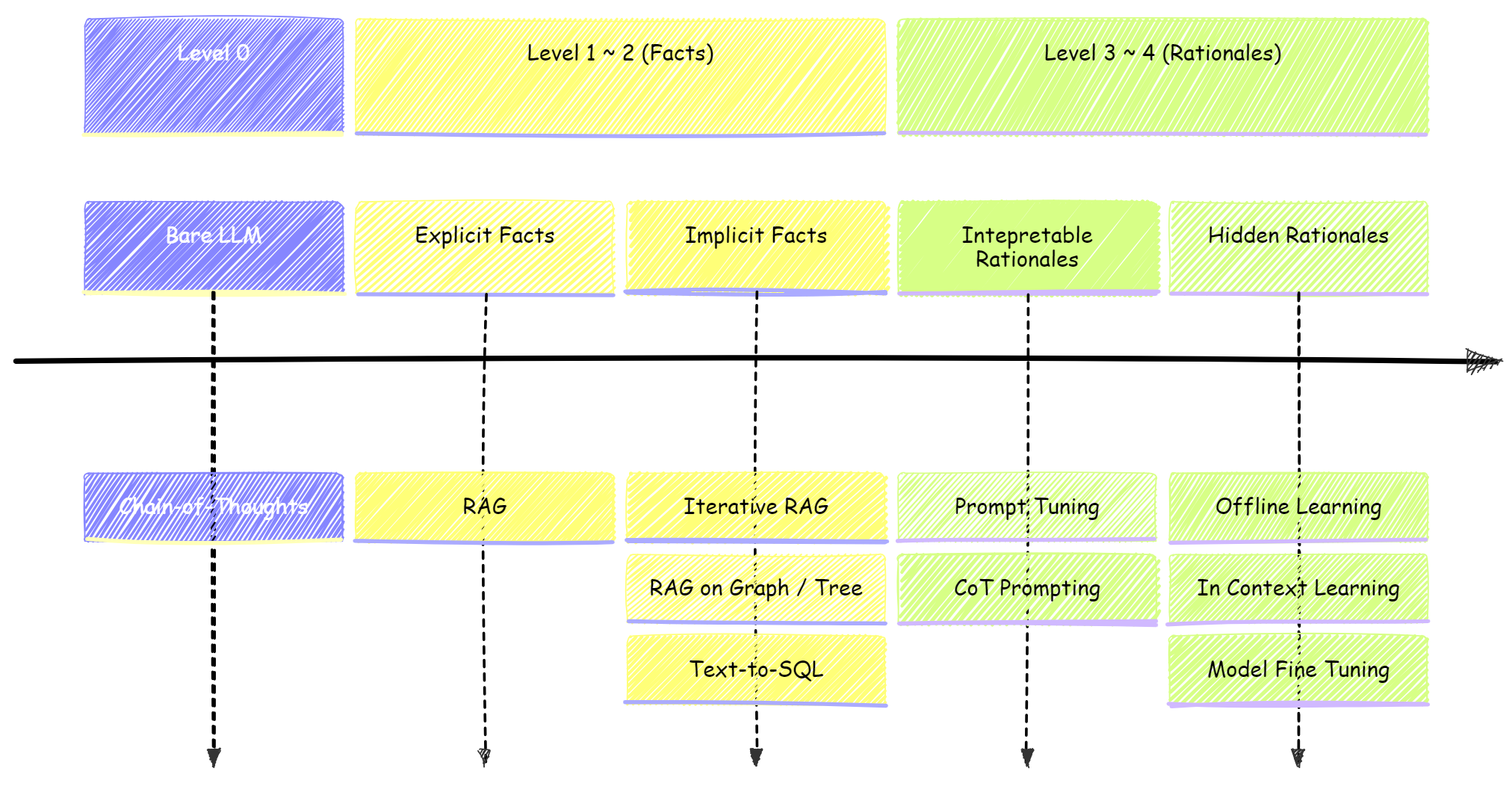}
    \caption{Summary of Main Techniques for Different Query Levels in \MFUsentencecase{\app}}
    \label{fig:conclusion}
\end{figure}

\subsection{Fine-tuning}

Despite the robust in-context learning capabilities of LLMs, accurately identifying rationales or optimal examples for complex and lengthy logical chains remains a significant challenge. Additionally, the provision of extensive external prior knowledge can also pose challenges to the inference capabilities of LLMs. Given these factors, fine-tuning emerges as a promising approach. It not only utilizes the extensive foundational knowledge that LLMs acquire during pretraining but also enables them to rapidly grasp new domain rationales. This method provides a viable path for enhancing the adaptability and effectiveness of LLMs in tackling advanced and specialized tasks.

Instruction tuning is a common method for infusing new capabilities into LLMs, typically involving supervised fine-tuning using paired (instruction, output) data. There are three primary methods for constructing an instruction dataset: a) deriving from existing datasets~\cite{longpre2023flan, sanh2021multitask}, b) manually creating through handcrafted instructions~\cite{ouyang2022training, conover2023free, kopf2024openassistant}, and c) generating synthetic data using powerful LLMs~\cite{xu2023wizardlm, wang2022self}. Additionally, numerous studies~\cite{raffel2020exploring, iyer2022opt, wang2023far} have explored how to optimize the data distribution within instruction datasets to enhance fine-tuning effectiveness. However, when building data-augmented LLM applications, fine-tuning remains a relatively costly method in terms of time and computational resources. Recently, several efforts have been made to reduce the costs associated with fine-tuning large models. Adapter tuning, for instance, involves integrating small adapter models with LLMs while freezing the parameters of the LLM during fine-tuning and only optimizing the weights of the adapter~\cite{houlsby2019parameter, he2021towards, lei2023conditional, chronopoulou2023adaptersoup}. Prefix Tuning and Prompt Tuning involve adding a set of trainable vectors before the input, which are optimized during training to enhance the performance of the LLM~\cite{petrov2023prompting, mahabadi2021parameter, zhu2023spt, wang2023aprompt, choi2023smop}. Low-Rank Adaptation~\cite{hu2021lora, valipour2022dylora, zhang2023adalora, ding2023sparse, liu2024dora} reduces the number of trainable parameters needed for adapting to downstream tasks by imposing low-rank constraints on each dense layer to approximate the update matrices.

In recent years, there has been a substantial amount of work using supervised fine-tuning to enhance the capabilities of LLMs in specialized domains such as mathematical reasoning, finance, law, and healthcare~\cite{yu2022alert, zong2023solving, yang2024leandojo}. For instance, ChatTimeLlama~\cite{yuan2024back} introduced an interpretable time reasoning instruction tuning dataset and fine-tuned on LLaMA~\cite{touvron2023llama1} to significantly improve the model's complex temporal reasoning, future event prediction capabilities, and interpretability. LISA~\cite{lai2024lisa} leveraged a small set of segment data samples that involve reasoning to fine-tune the multimodal LLM LLaVA, which resulted in substantial improvements in reasoning segmentation capabilities. MAmmoTH~\cite{yue2023mammoth} ingeniously constructed a mathematical example dataset that uniquely combines Chain of Thought and Program of Thought reasoning, ensuring broad coverage across different mathematical domains and enhancing the LLM's ability to solve general mathematical problems. ReFT~\cite{trung2024reft} proposes a method for learning from multiple annotated reasoning paths corresponding to the same problem. It automatically samples numerous reasoning trajectories for a given mathematical problem, leveraging the correct answer to generate reward signals. ChatDoctor~\cite{li2023chatdoctor} utilized a large dataset of 100,000 patient-doctor dialogues from a widely-used online medical consultation platform to fine-tune LLaMA, significantly enhancing the model's ability to understand patient needs and provide effective recommendations. FinGPT~\cite{yang2023fingpt} developed an open-source LLM fine-tuned on financial data using automated data curation and lightweight, low-rank adaptation techniques. DISC-LawLLM~\cite{yue2023disc} created a supervised fine-tuning dataset for the Chinese judicial domain, fine-tuning LLMs to effectively serve a variety of users in different legal scenarios with enhanced legal reasoning capabilities.
\section{Conclusion}

In this paper, we delineate \app{} into four distinct categories based on the primary focus of queries, each facing unique challenges and thus requiring tailored solutions, as illustrated in Figure~\ref{fig:conclusion}. For queries related to static common knowledge, deploying a general LLM through a Chain of Thought methodology is effective. For \firstlevel, the main challenge involves pinpointing the exact location of facts within a database, thus making a basic RAG the method of choice. In the case of \secondlevel, which require the collation of multiple related facts, iterative RAG and RAG implementations on graph or tree structures are preferred for their ability to concurrently retrieve individual facts and interconnect multiple data points. When extensive data linkage is necessary, text-to-SQL techniques prove indispensable, leveraging database tools to facilitate external data searches. For \thirdlevel, advancements through prompt tuning and CoT prompting are critical to enhance LLMs' compliance with external directives. The most formidable are \forthlevel, which demand the autonomous synthesis of problem-solving approaches from extensive data sets. Here, offline learning, in-context learning , and fine-tuning emerge as vital methodologies.

\begin{figure}
    \centering
    \includegraphics[width=0.4\textwidth]{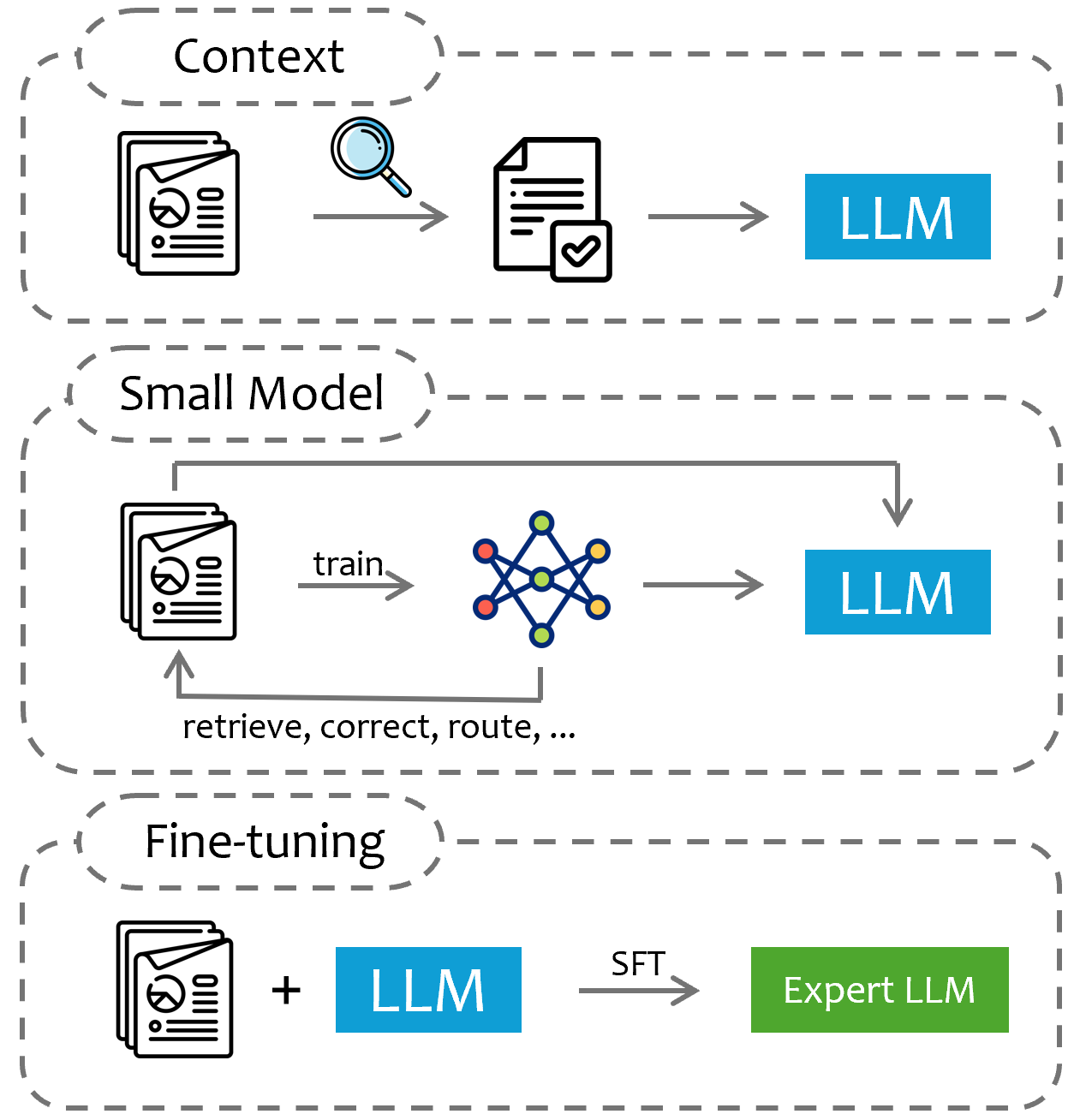}
    \caption{Three ways to inject specific domain data into an LLM: a) extracting part of the domain data based on the query as context input for the LLM, b) training a smaller model with specific domain data, which then guides the integration of external information subsequently input into the LLM, and c) directly using external domain knowledge to fine-tune a general large language model to become a domain-expert model.}
    \label{fig:three_type_input}
\end{figure}

Prior to the development of a targeted LLM application, as domain experts, we must acquire an in-depth understanding of the intended task, ascertain the complexity level of the associated queries, and select corresponding technological approaches for resolution. These methods principally inject knowledge into LLMs via three mechanisms, as depicted in Figure~\ref{fig:three_type_input}: a) extracting part of the domain data based on the query as context input for the LLM, b) training a smaller model with specific domain data, which then guides the integration of external information subsequently input into the LLM, and c) directly using external domain knowledge to fine-tune a general large language model to become a domain-expert model. These strategies differ in their requirements for data volume, training duration, and computational resources, escalating respectively. Knowledge injection through context provides better interpretability and stability but faces limitations due to the finite context window and potential information loss in the middle~\cite{liu2024lost}, ideally suited for scenarios where data can be succinctly explained in shorter texts. However, this method challenges the model's retrieval capabilities and knowledge extraction ability. The small model approach offers the advantage of reduced training times and the capacity to assimilate considerable amounts of data, yet its efficacy is contingent upon the model's capabilities, potentially capping the LLM’s performance for more complex tasks and incurring additional training costs with data increasing. Fine-tuning facilitates the utilization of large model capacities with extensive domain-specific data, yet its impact on the LLM strongly depends on the design of the data used. Employing out-of-domain factual data for fine-tuning may inadvertently lead to the generation of more erroneous outputs by the LLM, while also risking the loss of previously known domain knowledge and the neglect of unencountered tasks during fine-tuning~\cite{fine_tune_new_24, sun2024amuro}. Therefore, choosing an appropriate data injection strategy into the LLM requires a thorough understanding of one's data sources and judicious decision-making based on this insight.

Moreover, in practical scenarios, \app typically involves a combination of diverse query types, necessitating developers to engineer a routing pipeline that integrates multiple methodologies to effectively tackle these multifaceted challenges.

\bibliographystyle{unsrt}
\bibliography{contents/references}

\clearpage
\appendix

\end{document}